\pdfoutput=1

\documentclass[11pt]{article}

\usepackage[]{emnlp2021}

\usepackage{times}
\usepackage{latexsym}

\usepackage[T1]{fontenc}

\usepackage[utf8]{inputenc}

\usepackage{microtype}

\usepackage{booktabs} 
\usepackage{tikz}
\usepackage{fontawesome}
\usepackage{amsmath}
\usepackage{amssymb}
\usepackage{makecell}
\usepackage{tabularx}
\usepackage{xspace}
\usepackage{hyperref}
\usepackage{array}
\usepackage{subfig}
\usepackage{comment}
\usepackage{enumitem}

\DeclareMathOperator*{\argmax}{arg\,max}

\DeclareMathOperator*{\softmax}{softmax}

\newcommand\RobertaB{RoBERTa\textsubscript{Base}\xspace}
\newcommand\OurModel{MoLE\xspace}

\newcommand\DatasetsCount{16\xspace}

\newcommand\fmacro{F\textsubscript{1}}

\newcommand\dataurl{\url{https://github.com/checkstep/mole-stance}\xspace}

\newcolumntype{L}[1]{>{\raggedright\let\newline\\\arraybackslash\hspace{0pt}}m{#1}}
\newcolumntype{C}[1]{>{\centering\let\newline\\\arraybackslash\hspace{0pt}}m{#1}}
\newcolumntype{R}[1]{>{\raggedleft\let\newline\\\arraybackslash\hspace{0pt}}m{#1}}

\newcommand{\para}[1]{\noindent\textbf{#1}}

\title{Cross-Domain Label-Adaptive Stance Detection}

\author{
Momchil Hardalov$^{1,2}$ \quad
Arnav Arora$^{1,3}$ \quad
Preslav Nakov$^{1,4}$ \quad
Isabelle Augenstein$^{1,3}$ \\
$^1$Checkstep Research\\
$^2$Sofia University ``St. Kliment Ohridski'', Bulgaria\\
$^3$University of Copenhagen, Denmark\\
$^4$Qatar Computing Research Institute, HBKU, Doha, Qatar\\
{\tt \{momchil, arnav, preslav.nakov, isabelle\}@checkstep.com}
}

\begin{document}
\maketitle
\begin{abstract}
Stance detection concerns the classification of a writer's viewpoint towards a target. There are different task variants, e.g., stance of a tweet vs. a full article, or stance with respect to a claim vs. an (implicit) topic. Moreover, task definitions vary, which includes the label inventory, the data collection, and the annotation protocol. All these aspects hinder cross-domain studies, as they require changes to standard domain adaptation approaches. In this paper, we perform an in-depth analysis of \DatasetsCount stance detection datasets, and we explore the possibility for cross-domain learning from them. Moreover, we propose an end-to-end unsupervised framework for out-of-domain prediction of unseen, user-defined labels. In particular, we combine domain adaptation techniques such as mixture of experts and domain-adversarial training with label embeddings, and we demonstrate sizable performance gains over strong baselines, both (\emph{i})~in-domain, i.e., for seen targets, and (\emph{ii})~out-of-domain, i.e., for unseen targets. Finally, we perform an exhaustive analysis of the cross-domain results, and we highlight the important factors influencing the model performance. 
\end{abstract}

\section{Introduction}
\label{sec:introduction}

There are many different scenarios in which it is useful to study the attitude expressed in texts, e.g.,~of politicians with respect to newly proposed legislation~\citep{somasundaran-wiebe-2010-poldeb}, of customers regarding new products~\citep{somasundaran-wiebe-2009-recognizing}, or of the general public towards public health measures, e.g., aiming to reduce the spread of COVID-19~\citep{hossain-etal-2020-covidlies,glandt-etal-2021-stance}. This task, commonly referred to as \emph{stance detection}, has been studied in many different forms: not just for different domains, but with more substantial differences in the settings, e.g., stance (\emph{i})~expressed in tweets~\citep{qazvinian-etal-2011-rumor,mohammad-etal-2016-semeval,conforti-etal-2020-will-they} 
vs.~long news articles~\citep{pomerleau-2017-FNC,ferreira-vlachos-2016-emergent} vs. news outlets \cite{stefanov-etal-2020-user-stance} vs. people \cite{ICWSM2020:Unsupervised:Stance:Twitter},
(\emph{ii})~with respect to a claim~\citep{chen-etal-2019-perspectrum} vs. a topic, either explicit~\citep{qazvinian-etal-2011-rumor,walker-etal-2012-iac-corpus} or implicit~\citep{hasan-ng-2013-stance,gorrell-etal-2019-semeval}.
Moreover, there is substantial variation in (\emph{iii})~the label inventory, in the exact label definition, in the data collection, in the annotation setup, in the domain, etc. 
The most crucial of these, which has not been investigated, currently preventing cross-domain studies, is that the label inventories differ between the settings, as shown in Table~\ref{tab:dataset:domains}. Labels include not only variants of \emph{agree}, \emph{disagree}, and \emph{unrelated}, but also difficult to cross-map ones, such as \emph{discuss} and \emph{question}.

Our goal in this paper is to design a common stance detection framework to facilitate future work on the problem is a cross-domain setting. To this end, we make the following contributions:

\begin{itemize}
    \item We present the largest holistic study of stance detection to date, covering \DatasetsCount datasets.
    \item We propose a novel framework (\OurModel) that combines domain-adaptation and label embeddings for learning heterogeneous target labels.
    \item We further adapt the framework for out-of-domain predictions from a set of unseen targets, based on the label name similarity.
    \item Our proposed approach outperforms strong baselines both in-domain and out-of-domain.
    \item We perform an exhaustive analysis of cross-domain results, and find that the source domain, the vocabulary size, and the number of unique target labels are the most important factors for successful knowledge transfer.
\end{itemize}

Finally, we release our code, models, and data.\footnote{The datasets and code are available for research purposes:\\\dataurl}
\section{Related Work}
\label{sec:relatedwork}

\paragraph{Stance Detection}

Prior work on stance explored its connection to argument mining~\citep{Boltuzic2014Back-ur-Stance,sobhani2015-from-am-to-sc}, opinion mining~\citep{Wang2019-survey-opinion-mining}, and sentiment analysis~\citep{mohammad2017-stance-sentiment,aldayel-2019-stance-infer}.
Debating platforms were used as data source for stance~\citep{somasundaran-wiebe-2010-poldeb,Hasan2014WhyAY,aharoni-etal-2014-ibm-debater}, and more recently it was Twitter~\citep{mohammad-etal-2016-semeval,gorrell-etal-2019-semeval}.
With time, the definition of stance has become more nuanced 
\citep{kucuk-2020-stance-survey}, as well as its applications~\cite{zubiaga2018survey,hardalov2021survey}. Settings vary with respect to implicit~\citep{hasan-ng-2013-stance,gorrell-etal-2019-semeval} or explicit topics~\citep{augenstein-etal-2016-stance,stab-etal-2018-argmin,allaway-mckeown-2020-vast},
claims~\citep{baly-etal-2018-integrating,chen-etal-2019-perspectrum, hanselowski-etal-2019-snopes,conforti-etal-2020-stander,conforti-etal-2020-will-they}
or headlines~\citep{ferreira-vlachos-2016-emergent,habernal-etal-2018-arc,mohtarami-etal-2018-automatic}.

The focus, however, has been on homogeneous text, as opposed to cross-platform or cross-domain.
Exceptions are \citet{stab-etal-2018-argmin}, who worked on heterogeneous text, but limited to eight topics, and \citet{schiller2021stance}, who combined datasets from different domains, but used in-domain multi-task learning, and \citet{mohtarami-etal-2019-contrastive} and \citet{hardalov2021fewshot}, who used a cross-lingual setup.
In contrast, we focus on cross-domain learning on \DatasetsCount datasets, and out-of-domain evaluation.

\paragraph{Domain Adaptation}
Domain adaptation was studied in  \emph{supervised} settings, where in addition to the source-domain data, a (small) amount of labeled data in the target domain is also available~\citep{daume-iii-2007-frustratingly,finkel-manning-2009-hierarchical,donahue2013semi,yao2015semi,mou-etal-2016-transferable,lin-lu-2018-neural}, and in \emph{unsupervised} settings, without labeled target-domain data~\citep{blitzer-etal-2006-domain,lipton2018detecting,shah-etal-2018-adversarial,mohtarami-etal-2019-contrastive,Bjerva19-Future,wright-augenstein-2020-transformer}.
Recently, domain adaptation was applied to pre-trained Transformers \citep{lin-etal-2020-clinicalDA}.
One direction therein are architectural changes (method-centric):  \citet{ma-etal-2019-domain} proposed curriculum learning with domain-discriminative data selection, \citet{wright-augenstein-2020-transformer} investigated an unsupervised multi-source approach with Mixture of Experts and domain adversarial training~\citep{ganin-etal-2016-dann}. 

Another direction is data-centric adaptation:  \citet{han-eisenstein-2019-unsupervised,rietzler-etal-2020-adapt} used MLM fine-tuning on target-domain data.
\citet{gururangan-etal-2020-dont} showed alternate domain-adaptive (in-domain data) and task-adaptive (out-of-domain unlabelled data) pre-training. 

\paragraph{Label Embeddings} Label embeddings can capture, in an unsupervised fashion, the complex relations between target labels for multiple datasets or tasks. They can boost the end-task performance for various deep learning architectures, e.g.,~CNNs~\citep{zhang-etal-2018-multi,doi:10.1162/pappas_labelembed}, RNNs~\citep{augenstein-etal-2018-multi,augenstein-etal-2019-multifc}, and Transformers~\citep{chang2020taming}. Recent work has proposed different perspectives for learning label embeddings: \citet{beryozkin-etal-2019-joint} trained a named entity recogniser from heterogeneous tag sets,
\citet{chai2020description} used label descriptions for text classification,
\citet{rethmeier2021dataefficient} explored contrastive label embeddings for long-tail learning.

In our work, we propose an end-to-end framework to learn from heterogeneous labels based on unsupervised domain adaptation and label embeddings, and an unsupervised approach to obtain predictions for an unseen set of user-defined targets, using the similarity between label names.

\begin{table*}[t]
\centering
    \setlength{\tabcolsep}{3pt}
    \resizebox{1.00\textwidth}{!}{%
    \begin{tabular}{lllll}
    \toprule
    \bf Dataset & \bf Target & \bf Context & \bf Labels & \bf Source \\
    \midrule
    \makecell[l]{arc (\citeauthor{hanselowski-etal-2018-arc} \citeyear{hanselowski-etal-2018-arc};\\\citeauthor{habernal-etal-2018-arc} \citeyear{habernal-etal-2018-arc})} &               Headline &             User Post &                  unrelated (75\%), disagree (10\%), agree (9\%), discuss (6\%) &       Debates \\
    iac1~\citep{walker-etal-2012-iac-corpus}                     &                  Topic &       Debating Thread &                                          pro (55\%), anti (35\%), other (10\%) &       Debates \\
    perspectrum~\citep{chen-etal-2019-perspectrum}               &                  Claim &  Perspective Sent. &                                               support (52\%), undermine (48\%) &       Debates \\
    poldeb~\citep{somasundaran-wiebe-2010-poldeb}                &                  Topic &           Debate Post &                                                     for (56\%), against (44\%) &       Debates \\
    scd~\citep{hasan-ng-2013-stance}                             &  None (Topic) &           Debate Post &                                                     for (60\%), against (40\%) &       Debates \\
    \hline
    emergent~\citep{ferreira-vlachos-2016-emergent}              &               Headline &               Article &                                   for (48\%), observing (37\%), against (15\%) &          News \\
    fnc1~\citep{pomerleau-2017-FNC}                              &               Headline &               Article &                  unrelated (73\%), discuss (18\%), agree (7\%), disagree (2\%) &          News \\
    snopes~\citep{hanselowski-etal-2019-snopes}                  &                  Claim &               Article &                                                    agree (74\%), refute (26\%) &          News \\
    \hline
    mtsd~\citep{sobhani-etal-2017-mtsd-dataset}                  &                 Person &                 Tweet &                                      against (42\%), favor (35\%), none (23\%) &  Social Media \\
    rumor~\citep{qazvinian-etal-2011-rumor}                      &                  Topic &                 Tweet &  endorse (35\%), deny (32\%), unrelated (18\%), question (11\%), neutral (4\%) &  Social Media \\
    semeval2016t6~\citep{mohammad-etal-2016-semeval}             &                  Topic &                 Tweet &                                      against (51\%), none (24\%), favor (25\%) &  Social Media \\
    semeval2019t7~\citep{gorrell-etal-2019-semeval}              &  None (Topic) &                 Tweet &                        comment (72\%), support (14\%), query (7\%), deny (7\%) &  Social Media \\
    wtwt~\citep{conforti-etal-2020-will-they}                    &                  Claim &                 Tweet &                 comment (41\%), unrelated (38\%), support (13\%), refute (8\%) &  Social Media \\
    \hline
    argmin~\citep{stab-etal-2018-argmin}                         &                  Topic &              Sentence &                                   argument against (56\%), argument for (44\%) &       Various \\
    ibmcs~\citep{bar-haim-etal-2017-stance-ibmdebater}           &                  Topic &                 Claim &                                                         pro (55\%), con (45\%) &       Various \\
    vast~\citep{allaway-mckeown-2020-vast}                       &                  Topic &             User Post &                                         con (39\%), pro (37\%), neutral (23\%) &       Various \\
    \bottomrule
    \end{tabular}
    }
    \caption{List of the \DatasetsCount stance datasets we use and their characteristics (sorted by source, then alphabetically).}
    \label{tab:dataset:domains}
\end{table*}

\section{Stance Detection Datasets}
\label{sec:datasets}

In this section, we provide a brief overview of the \DatasetsCount stance datasets included in our study, and we show their key characteristics in Table~\ref{tab:dataset:domains}. More details are given in Section~\ref{subsec:datasets} and in the Appendix (Section~\ref{sec:appendix:data_splits}). We further motivate the source groupings used in our experiments and analysis (Section~\ref{subsec:sources}). 

\subsection{Datasets}
\label{subsec:datasets}

\para{arc} The Argument Reasoning Comprehension dataset has posts from the New York Times debate section on \textit{immigration} and \textit{international affairs}.

\para{argmin} The Argument Mining corpus presents arguments relevant to a particular topic from heterogenous texts. Topics include controversial keywords like \textit{death penalty} and \textit{gun control}.

\para{emergent} The Emergent\footnote{\label{emergent}\url{http://www.emergent.info/}} dataset is a collection of articles from rumour sites annotated by journalists.

\para{fnc1} The Fake News Challenge dataset consists of news articles whose stance towards headlines is provided. It spans 300 topics from Emergent.\footnotemark[2]

\para{iac1} The Internet Argument Corpus V1 consists of Quote--Response pairs from a debating forum on topics related to US politics.

\para{ibmcs} This dataset expands the IBM argumentative structure dataset \citep{aharoni-etal-2014-ibm-debater} to 55 topics and provides topic--claim pairs (from IBM Project Debater\footnote{IBM
Project Debater \url{http://www.research.ibm.com/artificial-intelligence/project-debater/}}) along with their stance annotations.

\para{mtsd} The Multi-Target Stance Detection dataset includes tweets related to the 2016 US Presidential election with a specific focus on multiple targets of interest expressed in each tweet.

\para{perspectrum} The Perspectrum dataset provides several perspectives towards a given claim gathered from a number of debating websites.

\para{poldeb} The Ideological On-Line Debates corpus provides opinion--target pairs from several debating platforms encapsulating different domains.

\para{rumor} The Rumor Has It dataset presents tweets for the task of Belief Classification, where users believe, question, or refute curated rumours.

\para{vast} The Varied Stance Topics dataset consists of topic--comment pairs from the The New York Times \textit{Room for Debate} section. The dataset covers a large variety of topics in order to facilitate zero-shot learning on new unseen topics.

\para{wtwt} Will-They-Won't-They presents a large number of annotated tweets from the financial domain relating to five merger and acquisition operations.

\para{scd} The Stance Classification dataset provides debate posts from four domains including \textit{Obama} and \textit{Gay Rights}. As highlighted in Table \ref{tab:dataset:stats}, while the posts are gathered from defined domains, they are not part of the training set and need to be inferred.\footnotetext[4]{\url{http://www.snopes.com}}

\para{semeval2016t6} The SemEval-2016 Task 6 dataset provides tweet--target pairs for 5 targets including \textit{Atheism}, \textit{Feminist Movement}, and \textit{Climate Change}.

\para{semeval2019t7} The SemEval-2019 Task 9 dataset aims to model authors' stance towards a particular rumour. It provides annotated tweets supporting, denying, querying, or commenting on the rumour.

\para{snopes} The Snopes dataset provides several controversial claims and their corresponding evidence texts from the US-based fact-checking website Snopes,\footnotemark{} annotated for the text in support of, refuting, or having no stance towards a claim. 

\subsection{Dataset Characteristics}
\label{subsec:dataset_characteristics}

\begin{table}[t]
    \centering
    \resizebox{1.00\columnwidth}{!}{%
    \begin{tabular}{l|rrrr} 
    \toprule
    \bf Dataset & \bf Train & \bf Dev &    \bf Test & \bf Total \\
    \midrule
    arc           &  12,382 &  1,851 &   3,559 &  17,792 \\
    argmin        &   6,845 &  1,568 &   2,726 &  11,139 \\
    emergent      &   1,770 &    301 &     524 &   2,595 \\
    fnc1          &  42,476 &  7,496 &  25,413 &  75,385 \\
    iac1          &   4,227 &    454 &     924 &   5,605 \\
    ibmcs         &     935 &    104 &   1,355 &   2,394 \\
    mtsd*         &   3,718  &   520 &   1,092 &   5,330 (8,910) \\
    perspectrum   &   6,978 &  2,071 &   2,773 &  11,822 \\
    poldeb        &   4,753 &  1,151 &   1,230 &   7,134 \\
    rumor*        &   6,093 &    471 &     505 &   7,276 (10,237) \\
    scd           &   3,251 &    624 &     964 &   4,839 \\
    semeval2016t6 &   2,497 &    417 &   1,249 &   4,163 \\
    semeval2019t7 &   5,217 &  1,485 &   1,827 &   8,529 \\
    snopes        &  14,416 &  1,868 &   3,154 &  19,438 \\
    vast          &  13,477 &  2,062 &   3,006 &  18,545 \\
    wtwt          &  25,193 &  7,897 &  18,194 &  51,284 \\
    \midrule        
    Total         & 154,228 & 30,547 & 68,495  & 253,270 \\ 
    \bottomrule
    \end{tabular}
    }
    \caption{Number of examples per data split. For datasets marked with \textsuperscript{*}, not all tweets could be downloaded; the original number of tweets is in parentheses.}
    \label{tab:dataset:stats}
\end{table}

\begin{figure}[t]
    \centering
    \includegraphics[width=1.0\columnwidth]{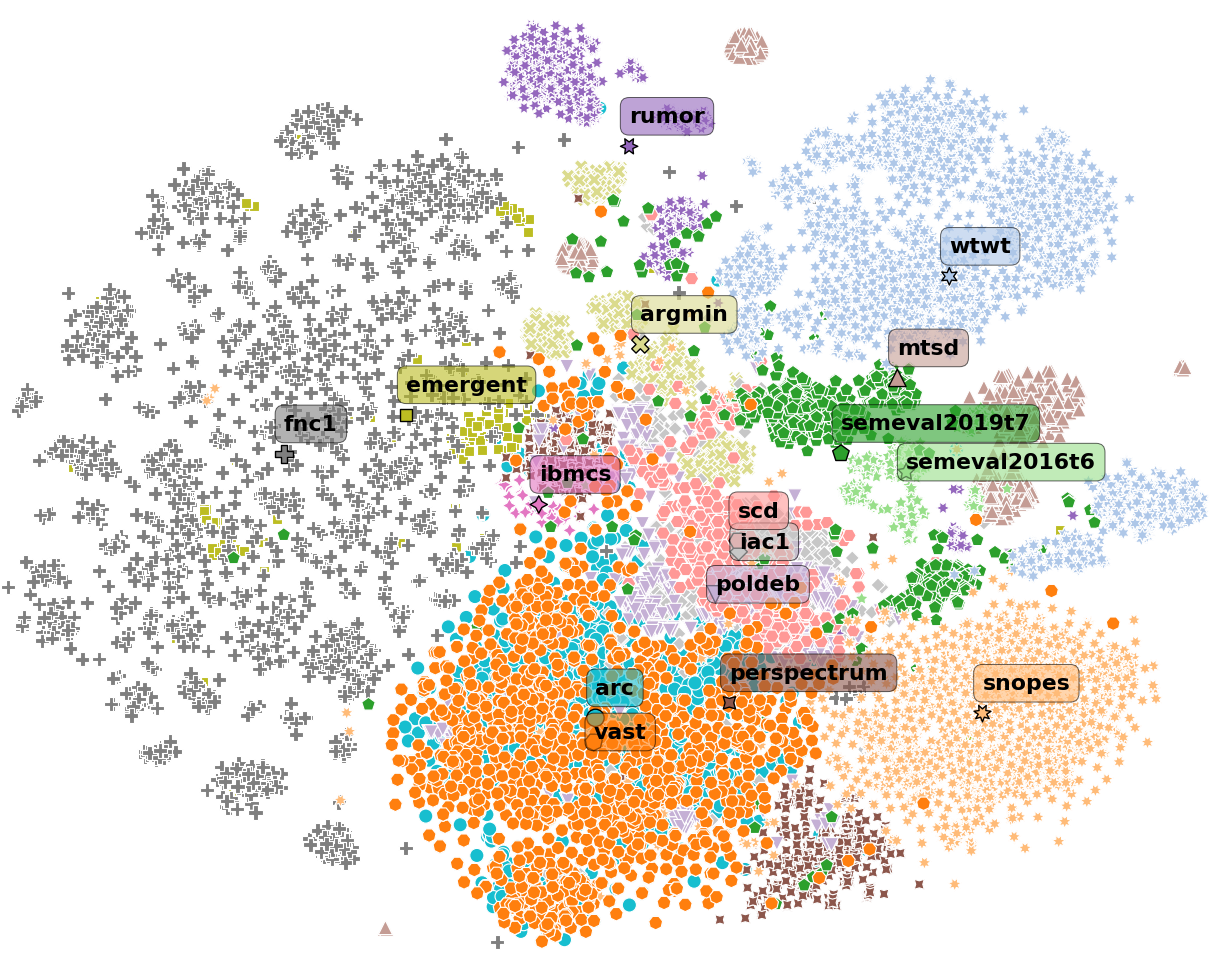}
    \caption{tSNE plot of \textit{[CLS]} representations of each dataset; highlighted points denote cluster centroids.}
    \label{fig:dataset:tsne_cls}
\end{figure}

As is readily apparent from Table~\ref{tab:dataset:domains}, the datasets differ based on the nature of the \emph{target} and the \emph{context}, as well as the stance \emph{labels}.

The \textit{Target} is the object of the stance. It can be a
\textit{Claim}, e.g.,~``\emph{Corporal punishment be used in K-12 schools.}'', 
a \textit{Headline}, e.g.,~``\emph{A meteorite landed in Nicaragua}'',
a \textit{Person}, 
a \textit{Topic}, e.g.,~\emph{abortion}, \emph{healthcare}, 
or \textit{None} (i.e.,~an implicit target).
Respectively, the \textit{Context}, which is where the stance is expressed, can be
an \textit{Article},
a \textit{Claim},
a \textit{Post}, e.g.,~in a debate, 
a \textit{Thread}, i.e.,~a chain of forum posts,
a \textit{Sentence},
or a \textit{Tweet}.
More examples from each dataset can be found in Table~\ref{tab:examples} in Appendix~\ref{sec:appendix:dataset}.

Moreover, the diversity of the datasets is also reflected in their label names, ranging from different variants of \textit{positive}, \textit{negative}, and \textit{neutral} to labels such as \emph{query} or \emph{comment}. The mapping between them is one of the core challenges we address, and it is discussed in more detail in Section~\ref{subsec:approach:label_adaptive}.

Finally, the datasets differ in their size (see Table~\ref{tab:dataset:stats}), varying from 800 to 75K examples. A complementary analysis of their quantitative characteristics, such as how the splits were chosen, the similarity between their training and testing parts, and their vocabularies, can be found in  Appendix~\ref{sec:appendix:dataset}.

\subsection{Source Groups}
\label{subsec:sources}

Defining source groups/domains is an important part of this study, as they allow for better understanding of the relationship between datasets, which we leverage through domain-adaptive modelling (Section~\ref{sec:approach}). Moreover, we use them to outline phenomena in the results that similar datasets share (Section~\ref{sec:experiments}). Table~\ref{tab:dataset:domains} shows these groupings.

Based on the aforementioned definitions of targets and context, we define the following groups: (\emph{i})~Debates, (\emph{ii})~News, (\emph{iii})~Social Media, and (\emph{iv})~Various.

We combine \textit{argmin} (Web searches), \textit{ibmcs} (Encyclopedia), and \textit{vast} into \emph{Various}, since they do not fit into any other group. 

To demonstrate the feasibility of our groupings, we plot the \DatasetsCount datasets in a latent vector space. We proportionally sample 25K examples, and we pass them through a \RobertaB~\citep{liu2019roberta} cased model without any training. The input has the following form: \texttt{[CLS]} context \texttt{[SEP]} target. Next, we take the \texttt{[CLS]} token representations, and we plot them in Figure~\ref{fig:dataset:tsne_cls} using tSNE~\citep{JMLR:v9:vandermaaten08a:tSNE}. We can see that Social Media datasets are grouped top-right, Debates are in the middle, and News are on the left (except for Snopes). The Various datasets, \textit{ibmcs} and \textit{argmin}, 
are placed in between the aforementioned groups (i.e.,~Debates and News), and \textit{argmin} is scattered into small clusters, confirming that they do not fit well into other source categories. Moreover, the figure reflects the strong connections between \textit{vast} and \textit{arc}, as well as between \textit{fnc1} and \textit{emergent}, as the former is derived from the latter. Finally, the clusters are well-separated and do not overlap, which highlights the rich diversity of the datasets, each of which has its own definition of stance.
\section{Method}
\label{sec:approach}

\begin{figure}[t]
    \centering
    \includegraphics[width=0.9\columnwidth]{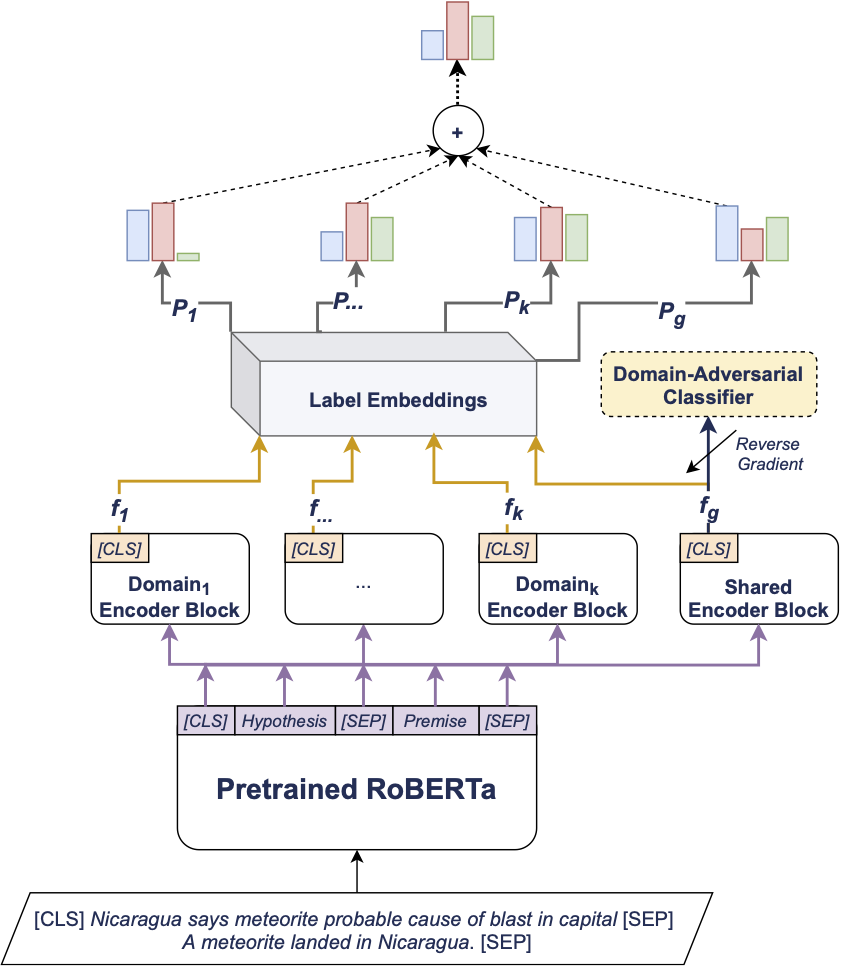}
    \caption{Overview of our proposed model \textit{\OurModel\ -- Mixture-of-Experts with Label Embeddings}.}
    \label{fig:approach:moe}
\end{figure}

We propose a novel end-to-end framework for cross-domain label-adaptive stance detection. Our architecture (see Figure~\ref{fig:approach:moe}) is based on input representations from a pre-trained language model, adapted to source domains using Mixture-of-Experts and domain adversarial training (Section~\ref{subsec:approach:proposed_model}). We further use self-adaptive output representations obtained via label embeddings, and unsupervised alignment between seen and unseen target labels for out-of-domain datasets.

Unlike previous work, we focus on learning the relationship between datasets and their label inventories in an unsupervised fashion. Moreover, our Mixture-of-Experts model is more compact than the one proposed by \citet{wright-augenstein-2020-transformer}, as we introduce a parameter-efficient architecture with layers that are shared between the experts. Finally, we explore the capability of the model to predict from unseen user-defined target.

With this framework, we solve two main challenges: (\emph{i})~training domain-adaptive models over a large number of datasets from a variety of source domains, and (\emph{ii})~predicting an \emph{unseen} label from a disjoint set of over 50 unique labels.

\subsection{Cross-Domain Stance Detection}
\label{subsec:approach:proposed_model}

\paragraph{Mixture-of-Experts (MoE)} is a well-known technique for multi-source domain adaptation~\citep{guo-etal-2018-multi,li-etal-2018-whats}. Recently, this framework was extended to large pre-trained Transformers~\citep{wright-augenstein-2020-transformer}.

In particular, for each domain $k \in K$, there is a domain expert $f_k$, and a shared, global model $f_g$.  We define $K=4$, as we use four different domains (see Section~\ref{subsec:sources}), making this approach appealing to further encourage knowledge sharing between datasets. Further, the models produce a set of probabilities $p_k$ from each expert and $p_g$ from the global model, for all the (often shared) target labels. Then, the final output of the model is obtained by passing these predictions through a combination function, e.g.,~mean average, weighted average, or attention-based combination~\citep{wright-augenstein-2020-transformer}. We use mean average to gather the final distribution across the label space:
\begin{equation}
    p_A(x, \bar{K}) = \frac{1}{|\bar{K}| + 1} \sum_{k \in \bar{K}} p_k(x) + p_g(x)
\end{equation}

\paragraph{Mixture-of-Experts with Label Embeddings (\OurModel)} We propose several changes to Mixture-of-Experts to improve the model's parameter efficiency, reduce training and inference times, and allow for different label inventories for each task.\footnote{In contrast, the model of~\citet{wright-augenstein-2020-transformer} requires that the datasets use the same labels.} First, in contrast to~\citet{wright-augenstein-2020-transformer}, we use a shared encoder, here, RoBERTa~\citep{liu2019roberta}, instead of a separate large Transformer model for each domain. Next, for each domain expert and the global shared model, we add a single Transformer~\citep{NIPS2017_7181:transformer} layer on top of the encoder block. We thereby retain the domain experts while sharing information through the encoder. This approach reduces the number of parameters by a factor of the size of the entire model divided by the size of a single layer, i.e.,~we only use four additional layers (one such encoder block per domain) instead of 48 (the number of layers in \RobertaB, not counting embedding layers). For convenience, we set the hidden sizes in the newly-added blocks to match the encoder's. 
Next, each domain expert receives as input the representations from the shared encoder of all tokens in the original sequence. Finally, we obtain a domain-specific and a global representation for the input sentence from the \texttt{[CLS]} tokens. These hidden representations are denoted as $H \in \mathbb{R}^{K \times d_h}$, where $K$ is the number of domains, and $d_h$ is the model's hidden size. They are passed through a single label embedding layer to obtain the probability distributions. 

\paragraph{Domain-Adversarial training} was introduced as part of the Domain-adversarial neural networks (\textbf{DANN})~\citep{pmlr-v37-ganin15}. 
The aim is to learn a task classifier by confusing an auxiliary domain classifier optimised to predict a meta target, i.e.,~the domain of an example. This approach has shown promising results for many NLP applications~\citep{li-etal-2018-whats,gui-etal-2017-part,wright-augenstein-2020-transformer}. Formally, it forces the model to learn domain-invariant representations, both for the source and for the target domains. The latter is done with an adversarial loss function, where we minimise the task objective $f_g$, and maximise the confusion in the domain classifier $f_d$ for an input sample $x$ (see Eq.~\ref{eq:approach:loss_dann}). We implement this with a gradient reversal layer, which ensures that the source and the target domains are made to be similar. 
\begin{equation}
    \mathcal{L}_D = \max_{\theta_{D}}\min_{\theta_{G}} -d\log{f_d(f_g(x))}
    \label{eq:approach:loss_dann}
\end{equation}

\subsection{Cross-Domain Label-Adaptive Prediction}
\label{subsec:approach:label_adaptive}

The second major challenge is how to obtain predictions for out-of-domain datasets. We want to emphasise that just a few of our \DatasetsCount datasets share the same set of labels (see Section~\ref{sec:datasets}); yet, many labels in different datasets are semantically related.

\paragraph{Label Embeddings (LEL)} In multi-task learning, each task typically has its own task-specific labels (in our case, dataset-specific labels), which are predicted in a joint model using separate output layers. 
However, these dataset-specific labels are not entirely orthogonal to each other (see Section~\ref{sec:datasets}). Therefore, we adopt label embeddings to encourage the model to learn task relations in an unsupervised fashion using a common vector space. In particular, we add a Label Embeddings Layer, or LEL, ~\cite{augenstein-etal-2018-multi}), which learns a label compatibility function between the hidden representation of the input $h$, here the one from the \texttt{[CLS]} token, and an embedding matrix $L$:
\begin{equation}
    p = \softmax(Lh)
\end{equation}
\noindent where $L \in \mathbb{R}^{(\sum_{i}{L_i}) \times h}$ is the shared label embedding matrix for all datasets, and $l$ is a hyper-parameter for the dimensionality of each vector. 

We set the size of the embeddings to match the hidden size of the model, and obtain the hidden representation $h$ from the last layer of the pre-trained language model. Afterwards, we optimise a cross-entropy objective over all labels, masking the unrelated ones and keeping visible only the ones from the target datasets for a sample in the batch. We use the same masking procedure at inference time.

\paragraph{Label-Adaptive Prediction} In an unsupervised out-of-domain setting, there is no direct way to obtain a probability distribution over the set of test labels.
Label embeddings are an easy indirect option for obtaining these predictions, as they can be used to measure the similarity between source and target labels.
We investigate several alternatives.

\para{Hard Mapping}
A supervised option is to
define a set of meta-groups (\emph{hard labels}), here six, as shown in Table~\ref{tab:approach:hard_map}, then to train the model on these labels. E.g., if the out-of-domain dataset is \emph{snopes}, then its labels are replaced with meta-group labels -- \emph{agree} $\Rightarrow$ \emph{positive}, and \emph{refute} $\Rightarrow$ \emph{negative}, and thus we can directly use the predictions from the model for out-of-domain datasets.
However, this approach has several shortcomings: (\emph{i})~labels have to be grouped manually, (\emph{ii})~the meta-groups should be large enough to cover different task definitions, e.g.~the dataset's label inventory may vary in size, and, most importantly, (\emph{iii})~any change in groupings would require full model re-training.

\begin{table}[t]
    \centering
    \setlength{\tabcolsep}{3pt}
    \resizebox{1.00\columnwidth}{!}{%
    \small
    \hyphenpenalty10000
    \begin{tabularx}{\columnwidth}{lL{0.8\columnwidth}}
    \toprule
    \bf Group & \bf Task\_\_Labels Included \\
    \midrule
    \makecell[l]{Positive} & arc\_\_\textit{agree}, argmin\_\_\textit{argument~for}, emergent\_\_\textit{for}, fnc1\_\_\textit{agree}, iac1\_\_\textit{pro}, mtsd\_\_\textit{favor}, perspectrum\_\_\textit{support}, poldeb\_\_\textit{for}, rumor\_\_\textit{endorse}, scd\_\_\textit{for}, semeval2016t6\_\_\textit{favor}, semeval2019t7\_\_\textit{support}, snopes\_\_\textit{agree}, vast\_\_\textit{pro}, wtwt\_\_\textit{support} \\
    \midrule
    \makecell[l]{Negative} &  arc\_\_\textit{disagree}, argmin\_\_\textit{argument~against}, emergent\_\_\textit{against}, fnc1\_\_\textit{disagree}, iac1\_\_\textit{anti}, ibmcs\_\_\textit{con}, mtsd\_\_\textit{against}, perspectrum\_\_\textit{undermine}, poldeb\_\_\textit{against}, rumor\_\_\textit{deny}, scd\_\_\textit{against}, semeval2016t6\_\_\textit{against}, semeval2019t7\_\_\textit{deny}, snopes\_\_\textit{refute}, vast\_\_\textit{con}, wtwt\_\_\textit{refute} \\
    \midrule
    \makecell[l]{Discuss}  & arc\_\_\textit{discuss}, emergent\_\_\textit{observing}, fnc1\_\_\textit{discuss}, rumor\_\_\textit{question}, semeval2019t7\_\_\textit{query}, wtwt\_\_\textit{comment} \\
    \midrule
    \makecell[l]{Other}    & arc\_\_\textit{unrelated}, fnc1\_\_\textit{unrelated}, iac1\_\_\textit{other}, mtsd\_\_\textit{none}, rumor\_\_\textit{unrelated}, semeval2019t7\_\_\textit{comment}, wtwt\_\_\textit{unrelated} \\
    \midrule
    \makecell[l]{Neutral}  & rumor\_\_\textit{neutral}, vast\_\_\textit{neutral} \\
    \bottomrule
    \end{tabularx}
    }
    \caption{Hard mapping of labels to categories.}
    \label{tab:approach:hard_map}
\end{table}

\para{Soft Mapping} To overcome these limitations, we propose a simple, yet effective, entirely unsupervised procedure involving only the label names. More precisely, we measure the similarity between the names of the labels across datasets. This is an intuitive approach for finding a matching label without further context, e.g.,~\textit{for} is probably close to \textit{agree}, and \textit{refute} is close to \textit{against}. 
In particular, given a set of out-of-domain target labels $Y^{\tau} \in \{y^{\tau}_1, \dots, y^{\tau}_k\}$, and a set of predictions from in-domain labels $P_{\delta} \in \{p^{\delta}_1, \dots, p^{\delta}_m\}, p^{\delta}_i \in \{y^{\delta}_1$, $\dots, y^{\delta}_j\}$, we select the label from $Y_{\tau}$ with the highest cosine similarity to the predicted label $p^{\delta}_i$:
\begin{equation}
    p^{\tau}_i = \argmax_{y^{\tau} \in Y^{\tau}} cos(y^{\tau}, p^{\delta}_i)
\end{equation}
where $k$ is the number of out-of-domain labels, $m$ the number of out-of-domain examples, and $j$ the number of in-domain labels. The procedure can generalise to any labels, without the need for additional supervision. To illustrate this, the embedding spaces of pre-trained embedding models for our \DatasetsCount datasets are visualised in Appendix~\ref{sec:appendix:lel_spaces}.

\para{Weak Mapping} Nevertheless, as proposed, this procedure only takes label names into account, in contrast to the hard labels that rely on human expertise. This makes combining the labels in a weakly supervised manner an appealing alternative. For this, we measure label similarities as proposed, but incorporate some supervision for defining the embeddings.
We first group the labels into six separate categories to define their nearest neighbours (see Table~\ref{tab:approach:hard_map}). Then, we choose the most similar label for the target domain from these neighbours.

The list of neighbours is defined by the group of the predicted label. However, there is no guarantee that there will be a match for the target domain within the same group, and thus we further define group-level neighbourhoods (see Table~\ref{tab:hand_neightbours}), as it is not feasible to define the neighbours for all (more than 50) labels individually.
One drawback is that each new label/group must define a neighbourhood with similar labels -- and vice-versa, it should be assigned a position in the neighbourhoods of the existing labels.

\begin{table}[t]
    \centering
    \small
    \begin{tabular}{l|l}
    \toprule
    \bf Group & \bf Closest Neighbours \\
    \midrule
    Positive & Other, Neutral, Discuss, Negative \\
    Other    & Neutral, Discuss, Positive, Negative \\
    Neutral  & Discuss, Other, Positive, Negative \\
    Discuss  & Neutral, Other, Negative, Positive \\
    Negative & Discuss, Neutral, Other, Positive \\
    \bottomrule
    \end{tabular}
    \caption{Label grouping and the closest neighbours of each, sorted from closest to most distant.}
    \label{tab:hand_neightbours}
\end{table}

\subsection{Training}

We train the model using the following loss:
\begin{gather}
    \mathcal{L}_s = \frac{1}{N}\sum_i{y_i\log{p_{X}(x, S')}} \\ 
    \mathcal{L}_t = \frac{1}{N}\sum_i{y_i\log{p_{t}(x)}} \\
    \mathcal{L} = \lambda\mathcal{L}_s + (1-\lambda)\mathcal{L}_t + \gamma\mathcal{L}_D
\end{gather}

First, we sum the source-domain loss ($\mathcal{L}_s$) with the meta-target loss from the domain expert sub-network ($\mathcal{L}_t$), where the contribution of each is balanced by a single hyper-parameter $\lambda$, set to $0.5$. Next, we add the domain adversarial loss ($\mathcal{L}_D$), and we multiply it by a weighting factor $\gamma$, which is set to a small positive number to prevent this regulariser from dominating the overall loss. We set $\gamma$ to $0.01$. Furthermore, since our dataset is quite diverse even in the four source domains that we outlined, we optimise the domain-adaptive loss towards a meta-class for each dataset, instead of the domain.

\section{Experiments}
\label{sec:experiments}

\begin{table*}[t]
    \centering
    \setlength{\tabcolsep}{3pt}
    \resizebox{1.00\textwidth}{!}{%
    \begin{tabular}{lc|ccccc|ccc|ccccc|ccc}
\toprule
{} &    \fmacro avg. &         arc &        iac1 & perspectrum &      poldeb &         scd &    emergent &        fnc1 &      snopes &        mtsd &       rumor & semeval16 & semeval19 &        wtwt &      argmin &       ibmcs &        vast \\
\midrule
Majority class baseline &     27.60 & 21.45 & 21.27 &        34.66 &   39.38 & 35.30 &     21.30 & 20.96 &   43.98 & 19.49 &  25.15 &          24.27 &          22.34 & 15.91 &   33.83 &  34.06 & 17.19 \\
Random baseline  &     35.19 & 18.50 & 30.66 &        50.06 &   48.67 & 50.08 &     31.83 & 18.64 &   45.49 & 33.15 &  20.43 &          31.11 &          17.02 & 20.01 &   49.94 &  50.08 & 33.25 \\
Logistic Regression & 41.35 & 21.43 & 28.68 & 61.33 & 72.30 & 44.63 & 61.30 & 24.02 & 59.32 & 44.29 & 19.31 & 48.92 & 22.34 & 32.32 & 51.06 & 37.13 & 33.31\\
\midrule
MTL w/  BERT\textsubscript{Base}           &       63.11 &       63.19 &  \bf{45.30} &       78.62 &       50.76 &       64.03 &  \bf{86.23} &       74.48 &       71.55 &       56.36 &       60.26 &         68.28 &    \bf{61.03} &       63.59 &       59.05 &       68.55 &       38.42 \\
MTL w/ \RobertaB         &       65.12 &       64.52 &       35.73 &       82.38 &       53.83 &       59.43 &       83.91 &       75.29 &       74.95 &  \bf{65.87} &  \bf{71.23} &         70.46 &    {59.42} &       67.64 &       61.79 &       77.27 &       38.21 \\
\midrule
\OurModel (Our Model)         &  \bf{65.55} &       63.17 &  {38.50} &  \bf{85.27} &       50.76 &  \bf{65.91} &       83.74 &       \bf{75.82} &  \bf{75.07} &      65.08 &       67.24 &         70.05 &         57.78 &       68.37 &  \bf{63.73} &       \bf{79.38} &  \bf{38.92} \\
$-$ DANN      &       65.40 &       64.28 &       37.20 &       83.93 &  \bf{53.99} &       62.79 &       83.44 &       75.47 &       74.77 &       65.44 &       70.41 &    \bf{72.08} &         54.68 &       68.90 &       62.29 &       78.42 &       38.24 \\
$-$ MoE    &       64.68 &  \bf{65.18} &       38.41 &       81.46 &       51.34 &       64.57 &  {84.60} &       75.79 &       74.05 &       65.69 &       61.07 &         69.99 &         56.67 &  \bf{69.03} &       62.25 &       76.87 &       37.91 \\

\bottomrule
\end{tabular}

    }
    \caption{\textbf{In-domain experiments.} Results are shown in terms of \fmacro. In the rows below \OurModel, we remove (\emph{--}) the components sequentially from it.}
    \label{tab:experiments:in_domain}
\end{table*}

\begin{table*}[t]
    \centering
    \setlength{\tabcolsep}{3pt}
    \resizebox{1.00\textwidth}{!}{%
    \begin{tabular}{lc|ccccc|ccc|ccccc|ccc}
\toprule
{} &    \fmacro avg. &         arc &        iac1 & perspectrum &      poldeb &         scd &    emergent &        fnc1 &      snopes &        mtsd &       rumor & semeval16 & semeval19 &        wtwt &      argmin &       ibmcs &        vast \\
\midrule
\OurModel w/ Hard Mapping &     32.78 & 25.29 & 35.15 &        29.55 &   22.80 & 16.13 &     58.49 & 47.05 &   29.28 & 23.34 &   32.93 &          \bf{37.01} &          21.85 & 16.10 &   34.16 &  72.93 & 22.89 \\
\OurModel w/ Weak Mapping              &       49.20 &  \bf{51.81} &  \bf{38.97} &       58.48 &       47.23 &  \bf{53.96} &  \bf{82.07} &       51.57 &  \bf{56.97} &  \bf{40.13} &  \bf{51.29} &         36.31 &    \bf{31.75} &       22.75 &       50.71 &  \bf{75.69} &  \bf{37.15} \\
\OurModel w/ Soft Mapping \\ 
w/ fasttext          &  \bf{42.67} &  \bf{48.31} &       13.23 &  \bf{62.73} &  \bf{54.19} &       49.58 &  \bf{46.86} &       53.46 &  \bf{53.58} &  \bf{37.88} &       44.38 &    \bf{36.77} &         24.40 &       21.53 &       56.48 &       59.26 &       19.67 \\
w/ glove             &       39.00 &       46.54 &        9.32 &       48.87 &       52.20 &  \bf{51.97} &       40.32 &       48.36 &       49.32 &       34.38 &  \bf{44.46} &         24.07 &          7.68 &  \bf{28.97} &  \bf{57.78} &       59.14 &       19.80 \\
w/ roberta-base      &       32.22 &       44.88 &  \bf{32.12} &       36.14 &       39.38 &       31.24 &       23.02 &       33.07 &       49.60 &       33.84 &       12.10 &         17.76 &          6.97 &       25.51 &       33.90 &       65.32 &  \bf{30.96} \\
w/ roberta-sentiment &       37.06 &       44.81 &       26.67 &       35.18 &       50.69 &       50.65 &       19.55 &       42.75 &       45.94 &       28.65 &       15.66 &         23.25 &    \bf{28.92} &       24.64 &       55.90 &  \bf{72.11} &       28.05 \\
w/ sswe              &       37.10 &       45.11 &       23.80 &       36.14 &       45.73 &       51.23 &       38.30 &  \bf{57.31} &       43.93 &       28.97 &       18.94 &         34.02 &          6.38 &       21.18 &       57.26 &       60.03 &       24.31 \\
\bottomrule
\end{tabular}
    }
    \caption{\textbf{Out-of-domain experiments.} Results are shown in terms of \fmacro.}
    \label{tab:experiments:out_domain}
\end{table*}

We consider three evaluation setups: (\emph{i})~\emph{no training}, random and majority class baselines; (\emph{ii})~\textit{in-domain}, training then testing on all datasets; and (\emph{iii})~\textit{out-of-domain}, i.e.,~leave-one-dataset-out training for all datasets. The reported per-dataset scores are
macro-averaged \fmacro,
which are additionally averaged to obtain per-experiment scores.

\subsection{Baselines}
\label{subsec:baselines}

\paragraph{Majority class baseline}
calculated from the distributions of the labels in each test set.

\paragraph{Random baseline} 
Each test instance is assigned a target label at random with equal probability.

\paragraph{Logistic Regression} A logistic regression trained using TF.IDF word unigrams. The input is a concatenation of the target and context vectors.

\paragraph{Multi-task learning (MTL)} A single projection layer for each dataset is added on top of a pre-trained language model (BERT~\citep{devlin2019bert} or RoBERTa~\citep{liu2019roberta}). We then pass the \texttt{[CLS]} token representations through the dataset-specific layer. Finally, we propagate the errors only through that layer (and the base model), without updating parameters for other datasets.

\subsection{Evaluation Results}
\label{subsec:evaluation_results}

\paragraph{In-Domain Experiments}  
We train and test on all datasets;
the results are in Table~\ref{tab:experiments:in_domain}. First, to find the best base model and set a baseline for \OurModel, we evaluate two strong models: BERT\textsubscript{Base}
uncased \citep{devlin2019bert}, and \RobertaB
cased\footnote{We choose the uncased version of BERT due to its wide use in similar tasks; RoBERTa is cased by nature. We use the base versions of the models for computational efficiency.} \citep{liu2019roberta}. On our \DatasetsCount datasets, RoBERTa outperforms BERT by 2 F\textsubscript{1} points absolute on average.

In the following rows of Table~\ref{tab:experiments:in_domain}, we show results for our model (\OurModel), i.e.,~Mixture of Experts with Label Embeddings and Domain-Adversarial Training (see Section~\ref{sec:approach}).
Its full version scores the highest in terms of \fmacro\ -- 65.55, which is 0.43 absolute points better than the MTL (\RobertaB) baseline. In particular, it outperforms this strong baseline on nine of the \DatasetsCount datasets. 
Nevertheless, neither \OurModel, nor any of its variations improves the results for \emph{mtsd}, \emph{rumor}, and \emph{semeval2019t6} over the MTL (\RobertaB) model. We attribute this to their specifics: \emph{mtsd} is the only dataset where the target is a \emph{Person}, \emph{rumor} and \emph{semeval2019t6} both focus on stance towards rumors, but the data for \emph{rumor} is from 2009--2011, and \emph{semeval2019t6} has an implicit target. 

Next, we present ablations -- we sequentially remove a prominent component from the proposed model (\OurModel). First, we optimise the model without the domain-adversarial loss. Removing the DANN leads to worse results on ten of the datasets, and a drop in the average \fmacro{} compared to \OurModel. However, this model does better in terms of points absolute on \emph{arc} (1\%), \emph{poldeb} (3\%), \emph{rumor} (3\%), and \emph{semeval2016t7} (+2\%). We attribute that to the more specialised domain representations being helpful, as some of the other datasets we trained on are very similar to those, e.g.,~\emph{vast} is derived from \emph{arc}.
Moreover, removing domain adversarial training has a negative impact on the datasets with source \emph{Various} (i.e.,~\emph{argmin, ibmcs, vast}).
Clearly, forcing similar representations aids knowledge sharing among domain experts, as they score between 0.7 and 1.5 \fmacro{} lower compared to \OurModel, the same behaviour as observed in other ablations.

The last row of Table~\ref{tab:experiments:in_domain} (\emph{$-$ MoE}) shows results for \RobertaB with Label Embeddings. It performs the worst of all RoBERTa-based models, scoring 0.5 points lower than MTL overall.
Note that it is not possible to present results for a MoE-based model without Label Embeddings, due to the discrepancy in the label inventories, both between and within domains, 
which means a standard voting procedure cannot be applied (see Section~\ref{subsec:approach:label_adaptive}).

\para{Out-of-Domain Experiments} 
In the out-of-domain setup, we leave one dataset out for testing, and we train on the rest.
We present results with the best model (\textbf{\OurModel}) on the in-domain setup as it outperforms other strong alternatives (see Table~\ref{tab:experiments:in_domain}).
In Table~\ref{tab:experiments:out_domain}, each column denotes when that dataset is held-out for training and instead evaluated on. 
We further evaluate all mapping procedures proposed in Section~\ref{subsec:approach:label_adaptive} for out-of-domain prediction: (\emph{i})~\emph{hard}
(\emph{ii})~\emph{weak}, 
and (\emph{iii})~\emph{soft mapping}. 

The \emph{hard mapping} approach outperforms the majority class baseline, but it falls almost 3 points absolute short compared to the random baseline, while failing to do better than random on more than half of the datasets. The two main factors for this are that (\emph{i})~the predictions are dominated by the meta-targets with the most examples, i.e.,~\emph{discuss}, (\emph{ii})~the model struggles to converge on the training set, due to diversity in the datasets and their labels. 

The \emph{weak} and the \emph{soft embeddings} share the same set of predictions, as their training procedure is the same -- the only difference between them are the embeddings used to align the prediction to the set of unseen targets. The \emph{weak mappings} achieve the highest average \fmacro{} among the out-of-domain models. For context, note that it is still 16\% behind the best in-domain model.
Furthermore, in this setup, we see that \emph{emergent} scores 82\%, just few points below the in-domain result -- we suspect that this is due to the good alignment of labels with \emph{fnc1}, as the two datasets are closely related.

For the \emph{soft mappings}, we evaluate five well-established embedding models, i.e.,~\emph{fastText}~\citep{joulin-etal-2017-bag}, \emph{GloVe}~\citep{pennington-etal-2014-glove}, \emph{\RobertaB}, and two sentiment-enriched ones, i.e.~\emph{sentiment-specific word embedding} (sswe,~\citet{tang-etal-2014-learning}), and \emph{RoBERTa Twitter Sentiment} (roberta-sentiment,~\citet{barbieri-etal-2020-tweeteval}). Our motivation for including the latter is that the names of the stance labels are often sentiment-related, and thus encoding that information into the latent space might yield better groupings (see Appendix~\ref{sec:appendix:lel_spaces}). 

We examine the performance of \emph{soft mapping w/ fastText} in more detail as they score the highest among other strong alternatives. Interestingly, the soft mappings benefit from splitting the predictions for the labels in the same group, such as \emph{wtwt\_\_comment} and all \emph{discuss-related}, which leads to the better performance on \emph{perspectrum, poldeb, fnc1, argmin} in comparison to the \emph{weak mappings}. Nevertheless, this also introduces some errors. An illustrative example are short words -- \emph{anti, pro, con}, which are distant from all other label names in our pool (see Figure~\ref{fig:lel_spaces:embeddings} in Appendix~\ref{sec:appendix:lel_spaces} for an illustration). 
The neighbourhoods are sometimes hard to interpret, e.g.,~\emph{con} is not the closest word for any predicted labels in \emph{vast}, and is aligned only with \emph{undermine, unrelated} in \emph{ibmcs}. 

\section{Discussion}
\label{sec:dicussions}

\begin{figure}[t!]
    \centering
    \includegraphics[width=0.87\columnwidth]{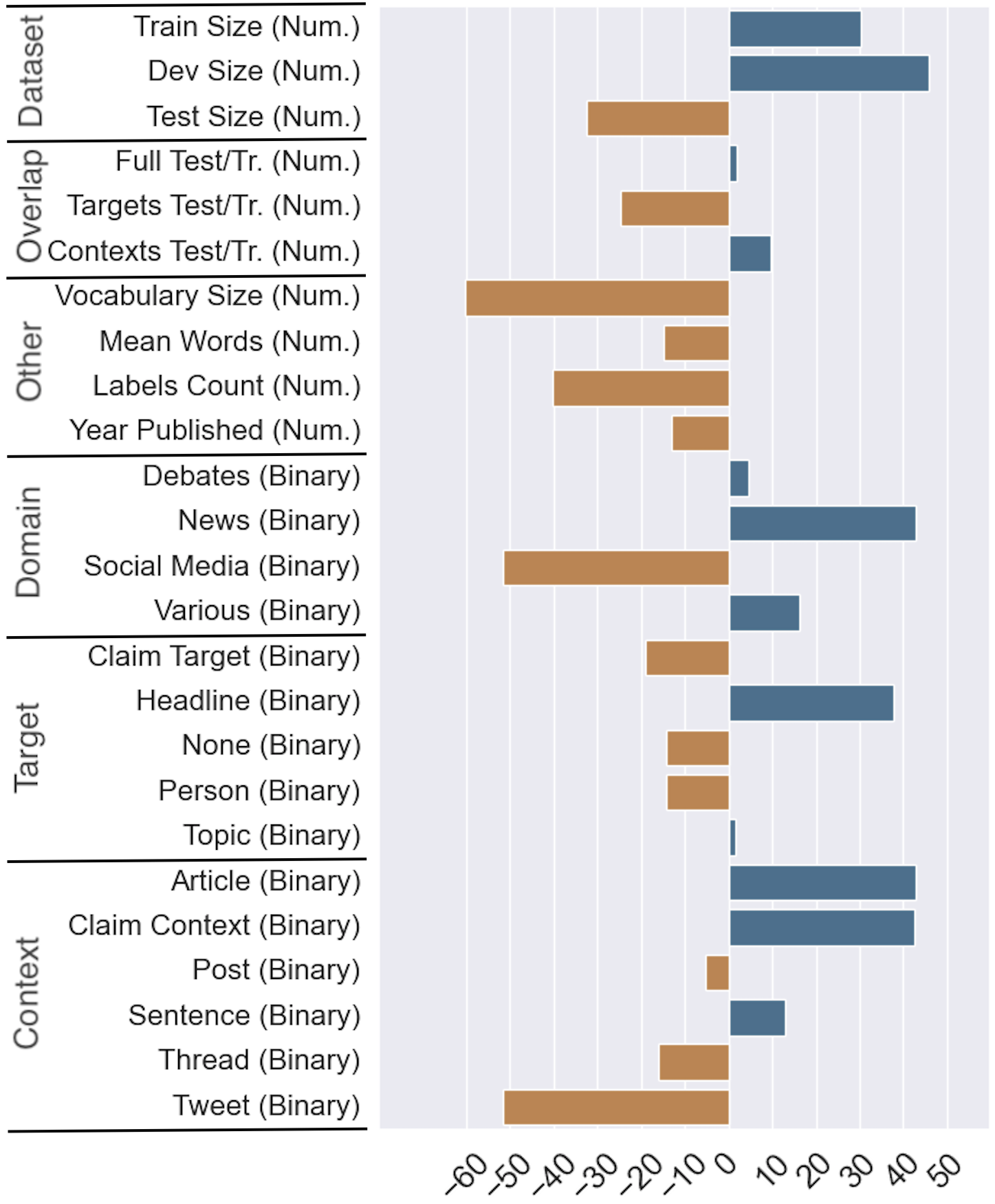}
    \caption{Pearson correlation between the out-of-domain results for our model (\OurModel w/ \emph{weak mappings}) in terms of \fmacro \ score and dataset characteristics. The (type) of the features is shown in parenthesis.}
    \label{fig:correlations}
\end{figure}

We further study the correlation between the scores for the best model in the out-of-domain setup \emph{\OurModel w/ weak mappings} and a rich set of quantitative and stance-related characteristics of the datasets (these are further discussed in Section~\ref{sec:datasets} and in Appendix~\ref{sec:appendix:dataset}). In particular, we represent each dataset as a set of features, e.g.,~\emph{fnc1} would have \emph{target -- News}, training set size of \emph{42,476}, etc., and then we measure the Pearson correlation between these features and the model's F\textsubscript{1} scores per dataset. 

Figure~\ref{fig:correlations} shows the most important factors for out-of-domain performance.\footnote{Some of the factors in the analysis are not independent, e.g., Social Media as a domain, and Tweet as a context.}
We see positive correlations of F\textsubscript{1} with the training, and the development set sizes, and a negative one with the testing set size, which suggests that large datasets are indeed harder for the model. Interestingly, if there is an overlap in the targets between the testing and the training sets, the model's \fmacro\ is worse; however, this is not true for context overlap. Unsurprisingly, the size of the vocabulary is a factor that negatively impacts F\textsubscript{1}, and its moderate negative correlation with the model's scores confirms that. 

The domain, the target and the context types are also important facets: the News domain has a sizable positive correlation with \fmacro, which is also true for the related features Headline target and Article body. Another positive correlation is for having a Claim as the context. On the contrary, a key factor that hinders model performance is Social Media text, i.e.,~having a tweet as a context. 

\section{Conclusion and Future Work}
\label{sec:conclusion}

We have proposed a novel end-to-end unsupervised framework for out-of-domain stance detection with respect to unseen labels. In particular, we combined domain adaptation techniques such as Mixture-of-Experts and domain-adversarial training with label embeddings, which yielded sizable performance gains on \DatasetsCount datasets over strong baselines: both in-domain, i.e.,~for seen targets, and out-of-domain, i.e., for unseen targets. Moreover, we performed an exhaustive analysis of the cross-domain results, and we highlighted the most important factors influencing the model performance. 

In future work, we plan to experiment with more datasets, including non-English ones, as well as with 
other formulations of the stance detection task, e.g., stance of a person \cite{ICWSM2020:Unsupervised:Stance:Twitter} or of a news medium \cite{stefanov-etal-2020-user-stance} with respect to a claim or a topic.

\section*{Acknowledgments}
We thank the anonymous reviewers for their helpful questions and comments, which have helped us improve the quality of the paper. 

We also would like to thank Guillaume Bouchard for the useful feedback. Finally, we thank the authors of the stance datasets for open-sourcing and providing us with their data.
\clearpage
\section*{Ethics and Broader Impact}

\paragraph{Dataset Collection}

We use publicly available datasets and we have no control over the way they were collected. For datasets that distributed their data as Twitter IDs, we used the Twitter API\footnote{\url{http://developer.twitter.com/en/docs}} to obtain the full text of the tweets, which is in accordance with the terms of use outlined by Twitter.\footnote{\url{http://developer.twitter.com/en/developer-terms/agreement-and-policy}} Note that we only downloaded public tweets.

\paragraph{Biases}

We note that some of the annotations are subjective. Thus, it is inevitable that there would be certain biases in the datasets. These biases, in turn, will likely be exacerbated by the supervised models trained on them~\citep{waseem2021disembodied}. This is beyond our control, as are the potential biases in pre-trained large-scale transformers such as BERT and RoBERTa, which we use in our experiments.

\paragraph{Intended Use and Misuse Potential}

Our models can enable analysis of text and social media content, which could be of interest to business, to fact-checkers, journalists, social media platforms, and policymakers.
However, they could also be misused by malicious actors, especially as most of the datasets we consider in this paper are obtained from social media. Most datasets compiled from social media present some risk of misuse. We, therefore, ask researchers to exercise caution.

\paragraph{Environmental Impact}
We would also like to note that the use of large-scale Transformers requires a lot of computations and the use of GPUs/TPUs for training, which contributes to global warming \cite{strubell-etal-2019-energy}. This is a bit less of an issue in our case, as we do not train such models from scratch; rather, we fine-tune them on relatively small datasets. Moreover, running on a CPU for inference, once the model has been fine-tuned, is perfectly feasible, and CPUs contribute much less to global warming.

\bibliographystyle{acl_natbib}
\bibliography{bibliography}

\clearpage
\appendix

\begin{table*}[tbh!]
    \centering
    \scriptsize
    \setlength{\tabcolsep}{3pt}
    \begin{tabularx}{\textwidth}{lp{3.5cm}Xl}
    \toprule
    Dataset & Target & Context & Label \\
    \midrule
    arc & States do not need special schools for the deaf & In the early 90's I was studying Maths of Finance at university and fees and charges were just starting to raise their ugly heads. In hindsight Australia is about 20 years ahead of the US with \dots & unrelated \\
    \midrule
    argmin & cloning & In Humanity Enhanced , I challenge the idea that children conceived through SCNT would have their autonomy violated – or would somehow lack or lose autonomy – in any sense inapplicable to “ ordinary ” children . & argument\_for \\
    \midrule
    emergent & Jess Smith of Chatham, Kent was the smiling sun baby in the Teletubbies TV show & Canterbury Christ Church University student Jess Smith, from Chatham, starred as Teletubbies sun & for \\
    \midrule
    fnc1 & Nigeria announces truce with Boko Haram; fate of schoolgirls unclear & Well, here’s the creepiest thing you’ll read all day. Australian Dylan Thomas found a tropical spider burrowed UNDER his skin after returning from a trip to Bali. \dots & unrelated \\
    \midrule
    iac1 & marijuana legalization & [P1] Instead of rewriting what I have already written numerous times, I will post a copy of a recent post I made on another thread regarding this same topic. You can find the original post here. \newline
    [P2] I enjoy the experience of getting drunk, high, or anything else like that...\dots & pro \\
    \midrule
    ibmcs & This house believes atheism is the only way & God is improbable & pro \\
    \midrule
    mtsd & Hilary Clinton & Love that \#democratic primary is talking bout real issues \#BernieSanders made a great case now let's hear from \#HillaryClinton \#DemTownHall & none \\
    \midrule
    perspectrum & The lack of investment in teachers is the greatest barrier to achieving universal primary education & It should be social policy to make teaching careers more desirable & support \\
    \midrule
    poldeb & should the us have universal healthcare & Yes, my posts all turned up backward because I replied in the order I read.Props for all the research you did. I still wholeheartedly disagree.Here's one article I came across that summarizes my entire point of view. http://url In the end, perhaps we agree to disagree \dots & against \\
    \midrule
    rumor & Sarah Palin getting divorced? & OneRiot.com - Palin Denies First Dude Divorce Rumors http://url & deny \\
    \midrule
    scd & -- & First off, the only people that want to legalize pot are the liberals that sit around all day, living off wellfare and smoke drugs. They do this because they are to uneducated to get a job, and figure the liberal \dots & against \\
    \midrule
    semeval2016t6 & Legalization of Abortion & @mrsdrjim did you know \#wrp @BrianJeanWRP tried to get personhood going via federal \#motion312. \#SemST & none \\
    \midrule
    semeval2019t7 & -- &  Wow, that is fascinating! I hope you never mock our proud Scandi heritage again. &  comment \\
    \midrule
    snopes & president-elect donald trump's inauguration will be the first presidential inauguration that rep. john lewis has skipped. & wrong (or lie)! & refute \\
    \midrule
    vast & public education & Public schools are the entire country's investment in an educated populace. They are our investment in a responsible, civil society. Everyone benefits when every citizen is able to read, write, understand history \dots & pro \\
    \midrule
    wtwt & Anthem acquires Cigna & - \#tuu i \#yoo - Anthem Reaffirms Commitment to Its \$47-Billion Bid for Cigna: Anthem stands by its \$47-bill... http://url & support \\
    \bottomrule
    \end{tabularx}
    \caption{Example from each dataset with stance target. The texts with \emph{...} are shortened due to space limitations.}
    \label{tab:examples}
\end{table*}

\section{Fine-Tuning and Hyper-Parameters}
\begin{itemize}
    \item All models are developed in Python using PyTorch~\citep{NEURIPS2019_9015} and the Transformers library~\citep{wolf-etal-2020-transformers}.
    \item All models use Adam~\citep{DBLP:journals/corr/KingmaB14} with weight decay 1e-8, $\beta_1$ 0.9, $\beta_2$ 0.999, $\epsilon$ 1e-08, and are trained for five epochs with batch size 64, and maximum length of 100 tokens.\footnote{When needed, we truncated the sequences token by token, starting from the longest sequence in the pair.}
    \item RoBERTa~\citep{liu2019roberta} is trained w/ LR 1e-05, warmup 0.06, BERT~\citep{devlin2019bert} is trained w/ LR 3e-05, warmup 0.1.
    \item The values of the hyper-parameters were selected on the development set.
    \item We chose the best model checkpoint based on the performance on the development set.
    \item For the MTL/MoE models, we sampled each batch from a single randomly selected dataset/domain.
    \item We used the same seed for all experiments.
    \item Each experiment took around 1h 15m on a single NVIDIA V100 GPU using half precision.
    \item For logistic regression, we converted the text to lowercase, removed the stop words, and limited the dictionary in the TF.IDF to 15,000 unigrams. We built the vocabulary using the concatenated target and context. The target and the context were transformed separately and concatenated to form the input vector. 
\end{itemize}

\section{Dataset Analysis}
\label{sec:appendix:dataset}

\subsection{Data Splits}
\label{sec:appendix:data_splits}

We could not reconstruct some of the Social Media datasets in full (marked with a \textsuperscript{*} symbol in Table~\ref{tab:dataset:stats}), as with only tweet IDs, we could not obtain the actual tweet text in some cases. This is a known phenomenon in Twitter: with time, older tweets become unavailable for various reasons, such as tweets/accounts being deleted or accounts being made private~\citep{zubiaga2018longitudinal}. The missing tweets were evenly distributed among the splits of the datasets except for \emph{rumor}, where we chose a topic for the test set for which all example texts were available.

Here, we provide more detail about the splits we used for the datasets, in cases where there is a deviation from the original. For the datasets in common, we used the splitting by \citet{schiller2021stance}. We further tried to enforce a larger domain diversity between the training, the development, and the testing sets; hereby, we put (whenever possible) all examples from a particular topic (domain) strictly into a single split.

\paragraph{argmin} We removed all non-arguments. The training, the development, and the test data splits consist of five, one, and two topics, respectively.

\paragraph{iac1} Split with no intersection of topics between the training, the development, and the testing sets.

\paragraph{ibmcs} Pre-defined training and testing splits. We further reserved 10\% of the training data for development set.

\paragraph{mtsd} We used the pre-defined splits, but we created two pairs for each example: a positive and a negative one with respect to the target.

\paragraph{poledb} We used the domains \textit{Healthcare, Guns, Gay Rights} and \textit{God} for training, \textit{Abortion} for development, and \textit{Creation} for testing.

\paragraph{rumor} We used the \textit{airfrance} rumour for our test set, and we split the remaining data in ratio 9:1 for training and development, respectively.

\paragraph{wtwt} We used \textit{DIS\_FOXA} operation for testing, \textit{AET\_HUM} for development, and the rest for training. To standardize the targets, we rewrote them as sentences, i.e.,~\emph{company X} acquires \emph{company Y}.

\paragraph{scd} We used a split with \textit{Marijuana} for development,  \textit{Obama} for testing, and the rest for training.

\paragraph{semeval2016t6} We split it to increase the size of the development set.

\paragraph{snopes} We adjusted the splits for compatibility with the stance setup. We further extracted and converted the rumours and their evidence into target--context pairs.

\begin{table}[t]
    \centering
    \resizebox{1.00\columnwidth}{!}{%
    \begin{tabular}{l|rrrrrr}
    \toprule
    {} & \multicolumn{3}{c}{\bf Dev} & \multicolumn{3}{c}{\bf Test} \\
    {\% of split in Train} & F &     T &    C & F &     T &    C \\
    \midrule
    arc           &  1.9 & 100.0 & 93.7 &  1.5 & 100.0 & 93.8 \\
    iac1          &  0.0 &   0.0 &  0.2 &  0.0 &   0.0 &  0.1 \\
    perspectrum   &  1.5 &   1.5 & 37.2 &  0.0 &   0.0 & 26.4 \\
    poldeb        &  0.0 &   0.0 &  0.0 &  0.0 &   0.0 &  0.0 \\
    scd           &   ---  &     --- &  0.0 &   --- &     --- &  0.0 \\
    emergent      &  0.0 &   0.0 &  3.0 &  0.0 &   0.0 &  1.7 \\
    fnc1          &  1.7 & 100.0 & 99.8 &  0.0 &   0.6 &  0.9 \\
    snopes        &  0.0 &   0.0 &  0.0 &  0.0 &   0.0 &  0.3 \\
    mtsd          &  0.8 & 100.0 &  1.5 &  0.3 & 100.0 &  0.5 \\
    rumor         & 17.6 & 100.0 & 17.6 &  0.0 &   0.0 &  0.0 \\
    semeval2016t6 &  0.0 & 100.0 &  0.0 &  0.0 & 100.0 &  0.0 \\
    semeval2019t7 &    --- &    ---  &  1.4 &    --- &    ---  &  4.8 \\
    wtwt          &  0.0 &   0.0 & 11.4 &  0.0 &   0.0 &  0.0 \\
    argmin        &  0.0 &   0.0 &  0.0 &  0.0 &   0.0 &  0.0 \\
    ibmcs         &  0.0 & 100.0 &  1.0 &  0.0 &   0.0 &  0.2 \\
    vast          &  0.0 &  43.6 &  0.0 &  0.0 &  49.5 &  0.0 \\
    \bottomrule
    \end{tabular}
    }
    \caption{Percentage of overlap  between development/testing and training data. The \textit{(T)arget} and the \textit{(C)ontext} columns show the overlap in the respective individual fields; \textit{(F)ull} shows the overall overlap.}
    \label{tab:dataset:overlaps}
\end{table}

\begin{table}[t]
    \centering
    \resizebox{1.00\columnwidth}{!}{%
    \begin{tabular}{l|rrrrr}
    \toprule
    {\bf Tokenization} & \multicolumn{1}{c}{\bf Words} & \multicolumn{1}{c}{\bf Tokens (Words)} & \multicolumn{3}{c}{\bf Tokens}  \\
    {} & \multicolumn{1}{c}{\bf $|$Unique$|$} & \multicolumn{1}{c}{\bf Mean}  & \bf 25\% & \bf Median & \bf Max \\
    \midrule
    arc           &     27,835 &    126.0 (118.3) &   84 &    116 &      286 \\
    argmin        &     17,990 &      30.6 (28.6) &   20 &     28 &      208 \\
    emergent      &      6,940 &      27.6 (23.1) &   22 &     26 &      111 \\
    fnc1          &     40,738 &    503.2 (432.2) &  279 &    413 &    6,182 \\
    iac1          &     88,478 &  1,554.7 (1,347.9) &  132 &    390 &  104,034 \\
    ibmcs         &      5,007 &      23.4 (21.9) &   18 &     23 &       55 \\
    mtsd          &      9,799 &      36.3 (25.9) &   33 &     36 &       65 \\
    perspectrum   &      9,999 &      22.4 (20.4) &   17 &     21 &       75 \\
    poldeb        &     40,422 &    178.0 (160.6) &   55 &    112 &    2,144 \\
    rumor         &     15,801 &      38.4 (24.4) &   32 &     39 &       78 \\
    scd           &     23,592 &    151.4 (134.8) &   39 &     78 &    6,358 \\
    semeval2016t6 &     15,016 &      32.8 (22.1) &   28 &     33 &       68 \\
    semeval2019t7 &     20,789 &      33.5 (22.0) &   17 &     27 &    1,466 \\
    snopes        &     33,896 &      53.4 (45.5) &   40 &     51 &      327 \\
    vast          &     24,644 &    123.2 (115.7) &   80 &    114 &      271 \\
    wtwt          &    102,672 &      45.7 (23.1) &   39 &     46 &      193 \\
    \bottomrule
    \end{tabular}
    }
    \caption{Statistics about the sub-word tokens for each dataset (using the \RobertaB tokeniser). The numbers in parenthesis show the word counts after the NLTK tokeniser was used.}
    \label{tab:dataset:vocab}
\end{table}

\subsection{Overlap Statistics}
\label{subsec:overlaps}

Next, in Table~\ref{tab:dataset:overlaps}, we examine the proportion of contexts and targets from the development and the testing datasets that are also present in the training split. We did not change the original data in any way, and we used the splits as described in Section~\ref{sec:appendix:data_splits}.

Table~\ref{tab:dataset:overlaps} further shows statistics about the datasets in terms of the number of words and sub-words they contain (see Table~\ref{tab:dataset:vocab}). The first column in the table shows the number of unique tokens (word types) in each dataset after tokenisation using NLTK's casual tokeniser~\citep{loper-2002-nltk}, which retains the casing of the words; thus word types of different casing are counted separately. We observe that the datasets with the largest vocabularies are those (\emph{i})~with higher numbers of examples (\emph{fnc1} and \emph{wtwt}), (\emph{ii})~whose contexts are threads rather than single posts (\emph{iac1} has over 1,300 words on average), and (\emph{iii})~that cover diverse topics such as 
\textit{poldeb} with six unrelated ones. In contrast, small or narrow datasets such as \textit{ibmcs} have the smallest vocabularies (fewer than 5,000 words). 

In the subsequent columns, we report statistics in terms of number of sub-words (i.e.,~SentencePieces~\citep{kudo-richardson-2018-sentencepiece} from \RobertaB's tokeniser). With that, we want to present the expected coverage in terms of tokens for a pre-trained model. On average, most of the datasets are well under 100 tokens in length, which is commonly observed for tweets,\footnote{Tweets have a strict character limit. Depending on the time period, this limit can vary.} but some datasets have a higher average number of tokens, e.g.,~debate-related datasets such as \textit{arc}, \textit{poldeb}, \textit{scd}, \textit{vast} fit in 200 tokens on average, which is also the case for datasets containing large news articles or use social media threads as context (\textit{fnc1}, \textit{iac1}), where the average length is over 500.

\begin{figure*}
    \centering
    \includegraphics[width=0.7\textwidth]{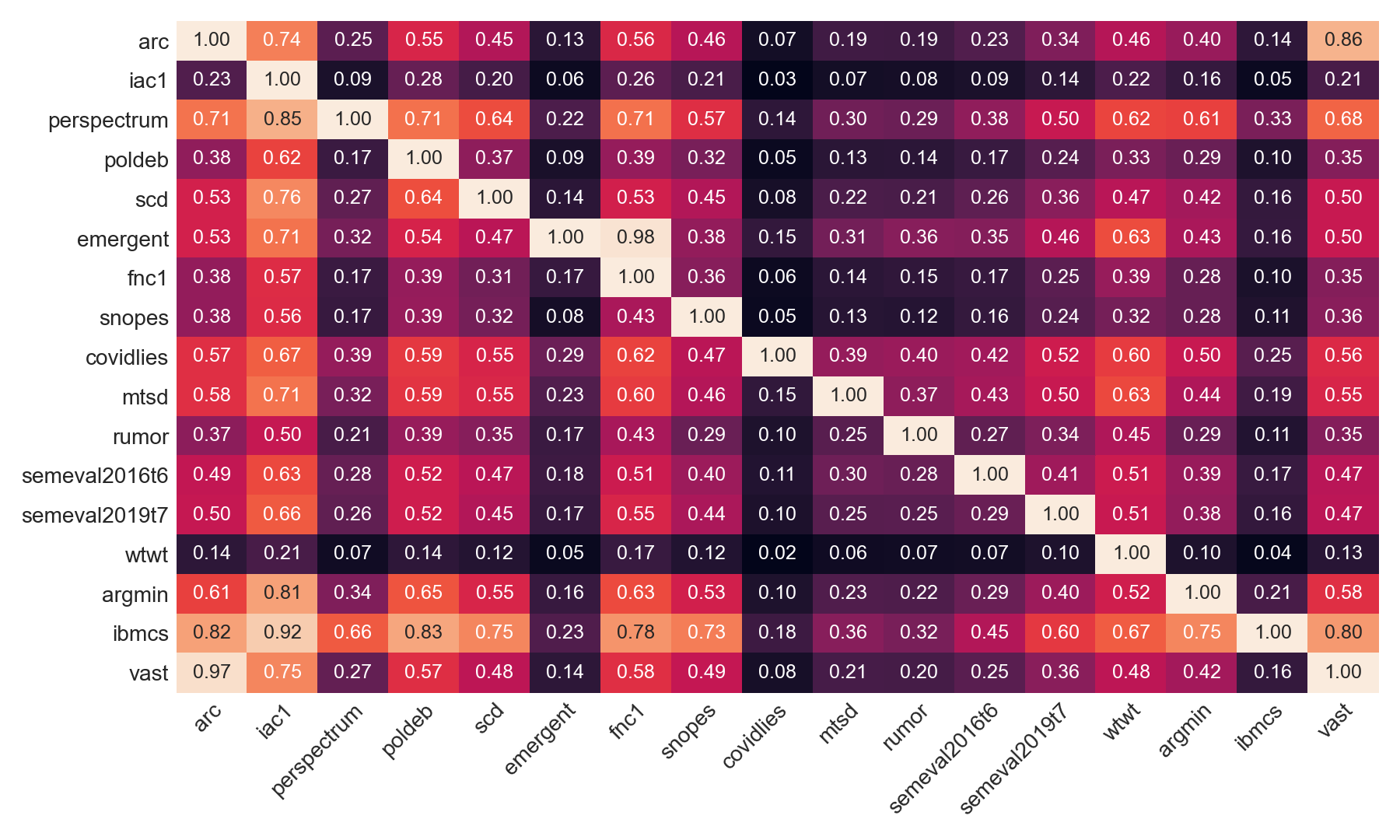}
    \caption{Proportion of word types in the dataset in row $i$ that are also present in the dataset in column $j$ (stop-words removed, and case preserved).}
    \label{fig:dataset:word_overlap}
\end{figure*}

Finally, Figure~\ref{fig:dataset:word_overlap} shows the relative word overlap between datasets. The numbers in each cell shows how much of the word types in dataset $i$ (row) are contained in the dataset $j$ (column). For example, in the first column in the last row (\textit{vast} $\Rightarrow$ \textit{arc}), we see that 97\% of the words in \textit{vast} are also present in \textit{arc}. Similarly, in the first row and the last column, we see that 86\% of the words in \textit{arc} are also in \textit{vast}. Note that we sort the columns and the rows by their sources (see Table~\ref{tab:dataset:domains}). We can see that datasets with the largest vocabularies (\textit{iac1} and \emph{wtwt}) have low overlap with other datasets, including with each other, up to 28\% only (row-wise).

When looking at how many words in other datasets are contained in them (column-wise), we see that \textit{iac1} has 50\% or more vocabulary overlap with the other datasets, even with ones from different sources. Then, \textit{wtwt}'s overlaps are 30--70\%, which is expected as its texts are from social media and cover a single topic (company acquisitions). For datasets that are either small or cover few topics (\emph{emergant, ibmcs, perspectrum}), we see that moderate to large part of their vocabularies is contained in other datasets; yet, the opposite in not true. Moreover, social media datasets are orthogonal to each other, with cross-overlaps of up to 50\% (both row-wise and column-wise), except for \emph{wtwt}. This is also seen when measuring how much of other datasets' vocabulary they contain (column-wise).

\begin{figure*}[th!]
    \centering
    \subfloat[FastText \label{subfig:embeddings:fasttext}]{
      \includegraphics[width=0.30\textwidth]{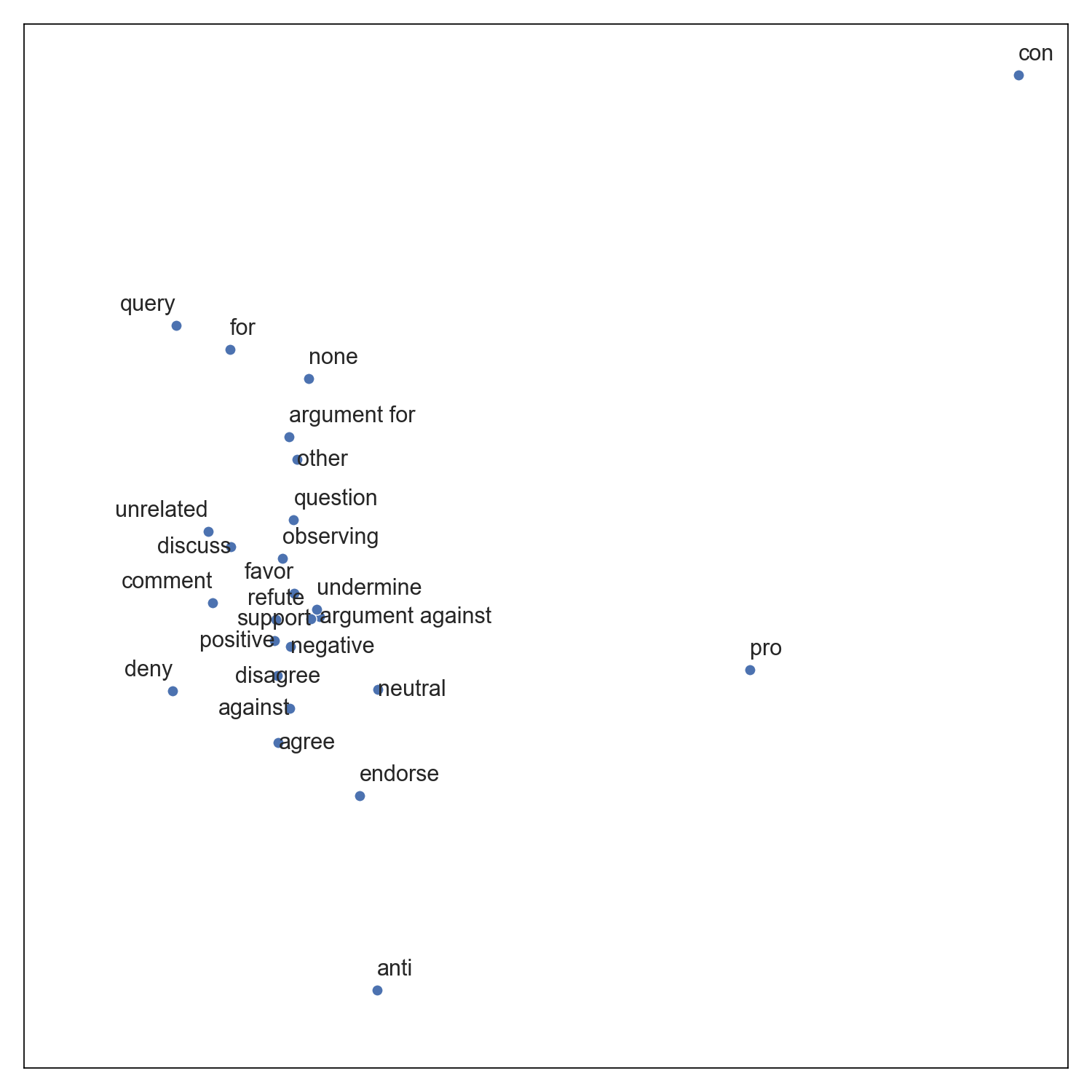}
    }
    \subfloat[GloVe \label{subfig:embeddings:glove}]{%
        \includegraphics[width=0.30\textwidth]{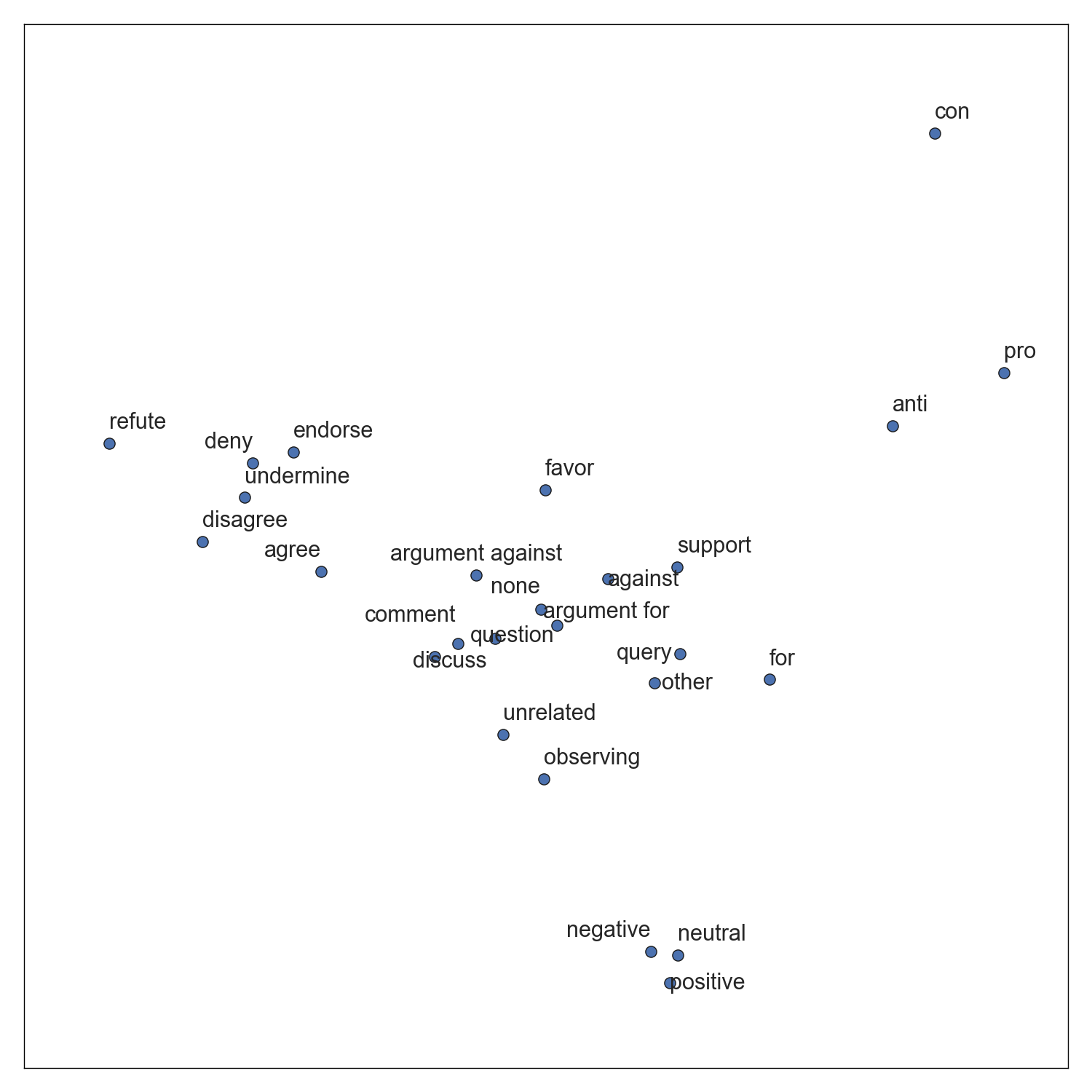}
    }
    \subfloat[RoBERTa-base \label{subfig:embeddings:roberta}]{
      \includegraphics[width=0.30\textwidth]{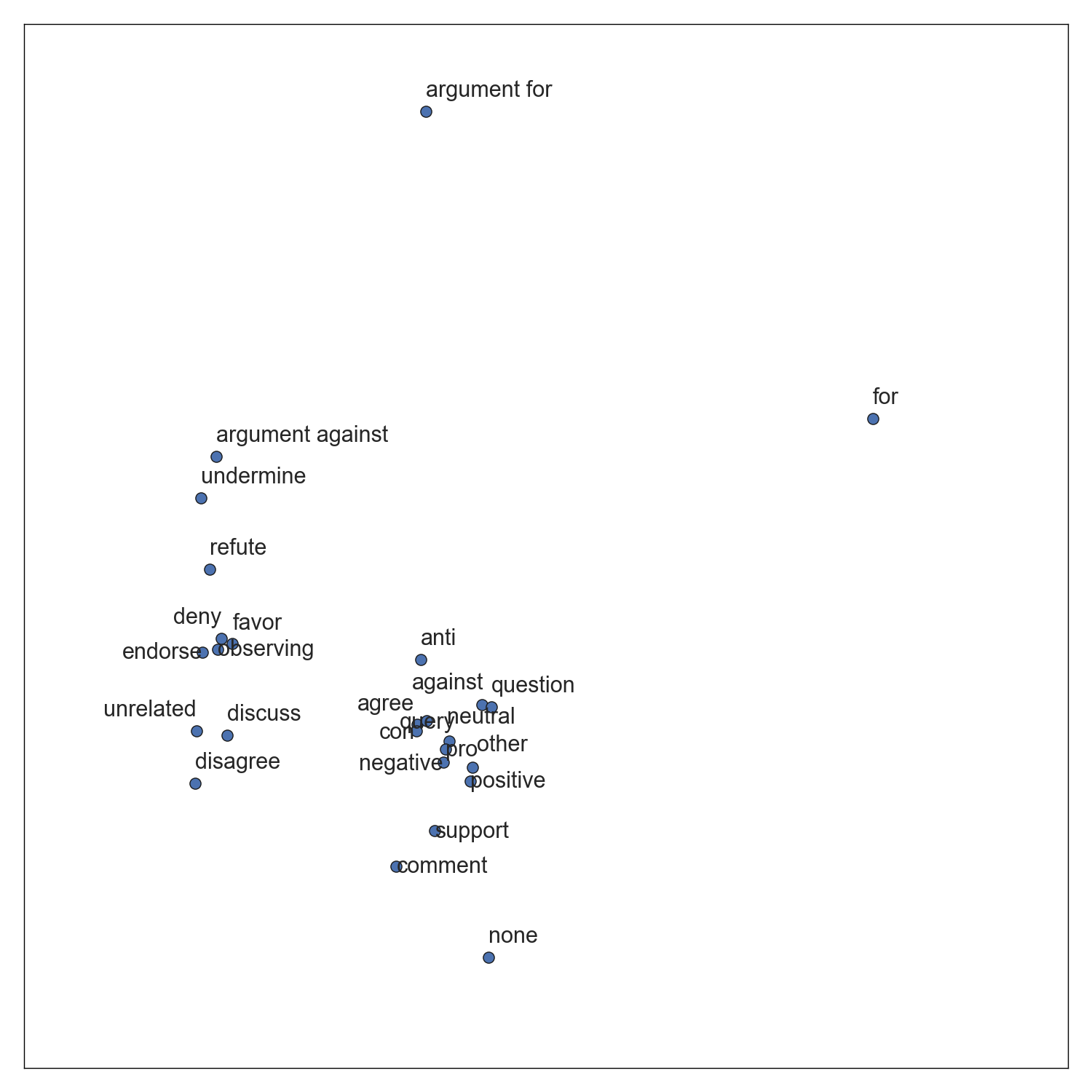}
    } \\
    \subfloat[RoBERTa-sentiment \label{subfig:embeddings:roberta_senti}]{
        \includegraphics[width=0.30\textwidth]{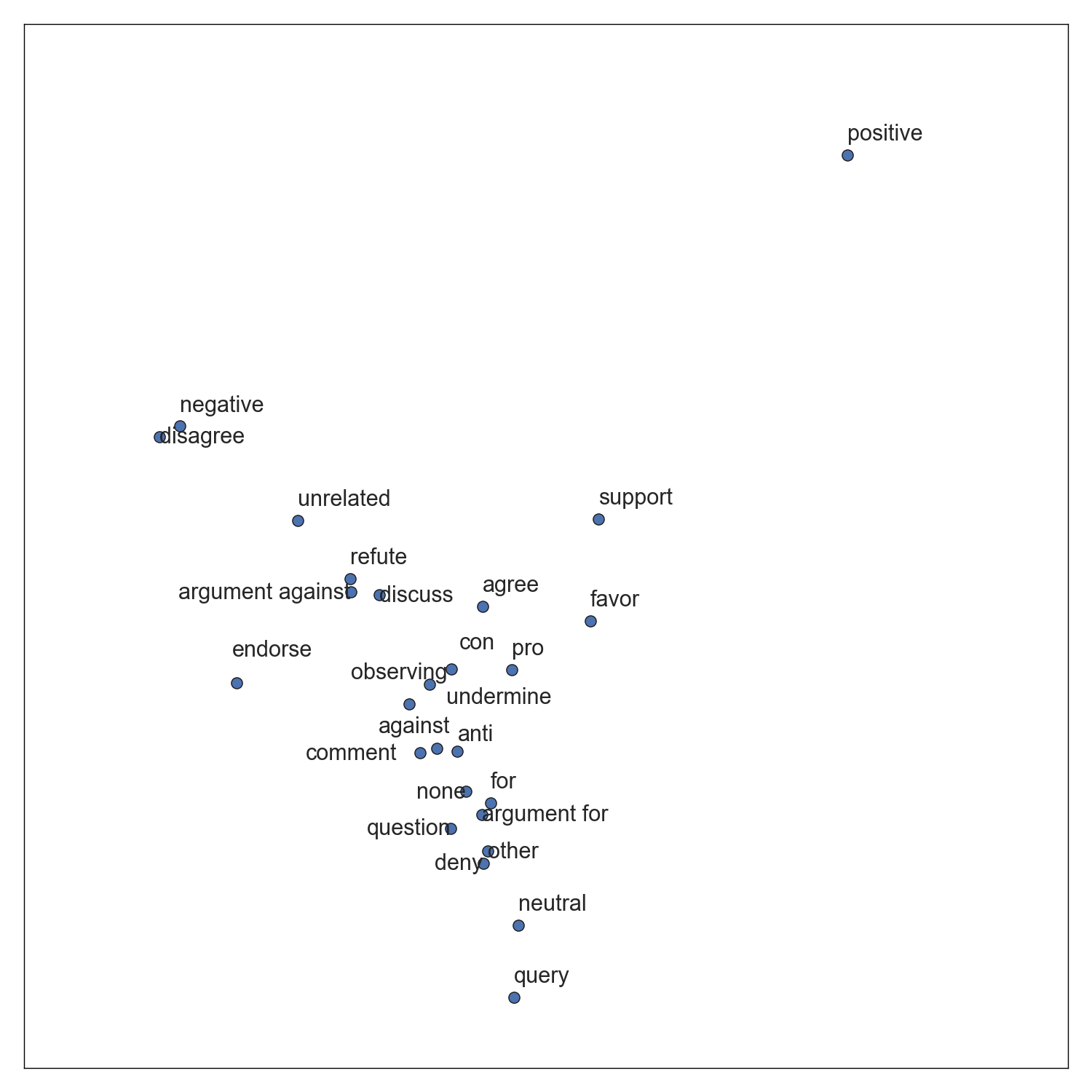}
    }
    \noindent\subfloat[SSWE \label{subfig:embeddings:sswe}]{%
        \includegraphics[width=0.30\textwidth]{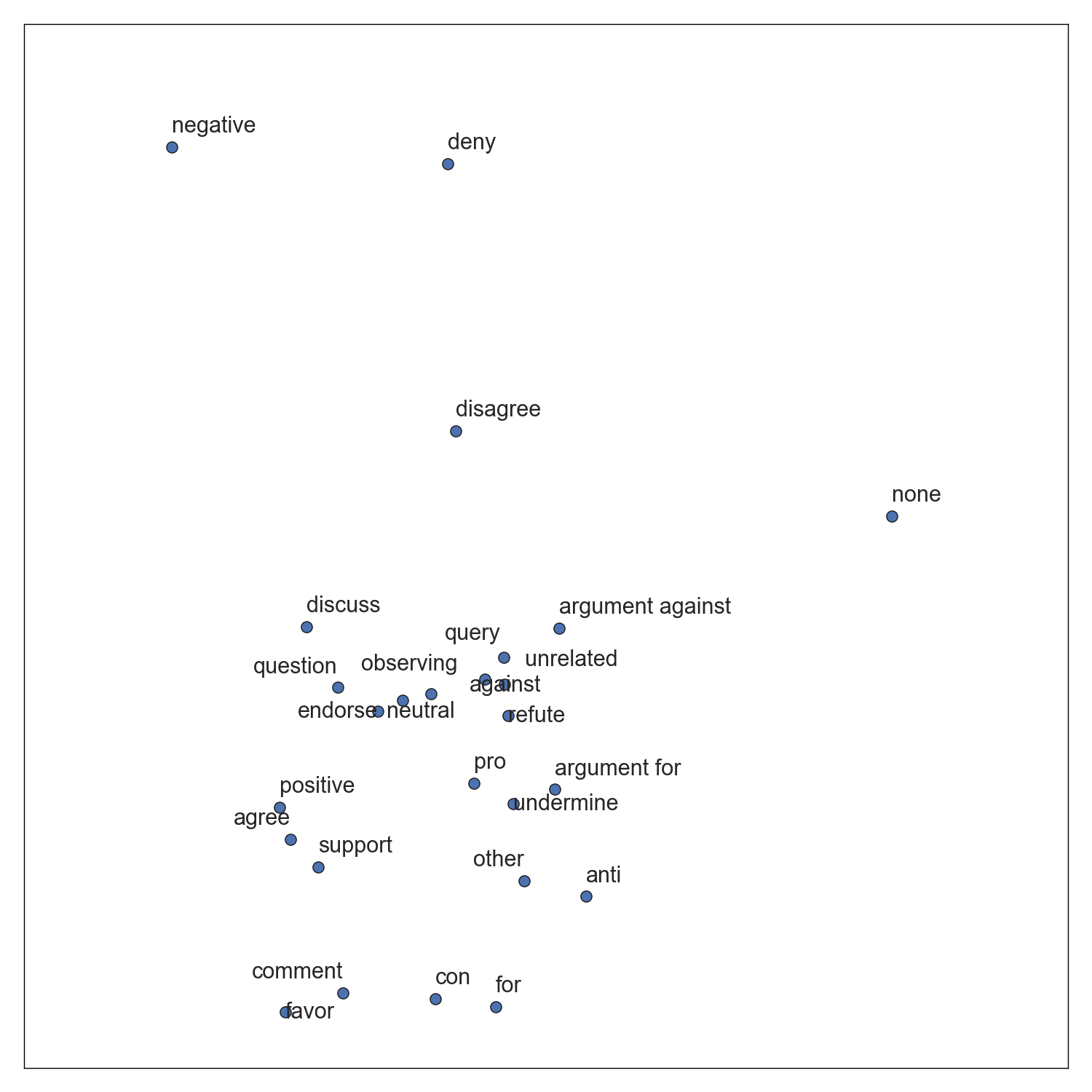}
    }
    \caption{Embedding spaces of the label names' representations (PCA) from different embedding models.}
    \label{fig:lel_spaces:embeddings}
\end{figure*}

\section{Embedding Label Spaces}
\label{sec:appendix:lel_spaces}

Here, we present the embeddings based on the labels' names. We explore five sets of embeddings: (\emph{i})~well-established ones such as \emph{fastText}~\citep{joulin-etal-2017-bag}, \emph{GloVe}~\citep{pennington-etal-2014-glove}, and \emph{RoBERTa}~\citep{liu2019roberta}, and (\emph{ii})~sentiment-enriched ones like \emph{Sentiment-specific word embedding} (SSWE)~\citep{tang-etal-2014-learning} and \emph{RoBERTa Twitter Sentiment} (roberta-sentiment)~\citep{barbieri-etal-2020-tweeteval}. Our motivation for including the latter is that the names of the stance labels are often sentiment-related, and thus encoding that information into the latent space could yield better grouping. We encode each label with the corresponding word from the embedding's directory for non-contextualized embeddings, if the name contains multiple words, e.g.,~\emph{argument for}, then we split on white space, and we take the average of each word's embeddings. For RoBERTa-based models, we take the representation from the \texttt{[CLS]} token. Finally, we project the obtained vectors into two dimensions using PCA. The resulting plots are shown in Figure~\ref{fig:lel_spaces:embeddings}. 

\end{document}